\ifdefined\pdfoutput\pdfoutput=1\fi
\documentclass[11pt]{article}

\usepackage[]{acl}

\usepackage{times}
\usepackage{latexsym}

\usepackage[T1]{fontenc}

\usepackage[utf8]{inputenc}

\usepackage{microtype}

\usepackage{inconsolata}

\usepackage{multirow, booktabs}
\usepackage{makecell, hhline}
\usepackage{xspace}
\usepackage{subfigure}
\usepackage{pifont}
\usepackage{longtable}
\usepackage[para,online,flushleft]{threeparttable}
\usepackage{algorithm}
\usepackage{algpseudocode}
\usepackage{amsmath,amssymb,mathtools,amsfonts,amsthm}

\usepackage{adjustbox}
\usepackage{algorithm}
\usepackage{algpseudocode}

\usepackage[most]{tcolorbox}
\usepackage{xcolor}
\usepackage{xurl}

\usepackage{siunitx} 
\definecolor{hookersgreen}{rgb}{0.0, 0.44, 0.0}
\definecolor{indiagreen}{rgb}{0.07, 0.53, 0.03}
\definecolor{islamicgreen}{rgb}{0.0, 0.56, 0.0}
\definecolor{kellygreen}{rgb}{0.3, 0.73, 0.09}
\definecolor{alizarin}{rgb}{0.82, 0.1, 0.26}

\algnewcommand\algorithmicinput{\textbf{Input:}}
\algnewcommand\algorithmicoutput{\textbf{Output:}}
\algnewcommand\algorithmicparameter{\textbf{Parameters:}}
\algnewcommand\INPUT{\item[\algorithmicinput]}
\algnewcommand\OUTPUT{\item[\algorithmicoutput]}
\algnewcommand\PARAMETER{\item[\algorithmicparameter]}

\newcommand{\ourmethod}{\textsc{LA-RL}}

\newcommand{\ExpTableStyle}{%
    \setlength{\tabcolsep}{8.5pt}%
}
\newcommand{\DenseExpTableStyle}{%
    \setlength{\tabcolsep}{4.2pt}%
}
\newcommand{\StageI}{LA-RL$^{\mathrm{I}}$}
\newcommand{\StageII}{LA-RL$^{\mathrm{II}}$}

\title{LA-RL: Label-Aware Self-Reflection for Reinforcement Learning in Information Extraction}

\author{
  Xiao You$^{1}$ \quad
  Tianwei Yan$^{2}$ \quad
  Zixu Shan$^{1}$ \quad
  Longyu Du$^{1}$ \quad
  Shan Zhao$^{1}$ \\
  $^{1}$Hefei University of Technology, Hefei, China \\
  $^{2}$Chongqing Jiaotong University, Chongqing, China
}

\begin{document}
\maketitle

\def\thefootnote{\arabic{footnote}}

\begin{abstract}
Large language models show strong promise for information extraction (IE), but existing reflection-based correction methods are often misaligned with structured extraction outputs. Free-form self-reflection can flag an error, yet it rarely identifies whether the failure is a missing span, wrong label, boundary mismatch, invalid relation type, or reversed argument order. We introduce LA-RL (Label-Aware Reflective Reinforcement Learning), an outcome-supervised framework that guides IE self-correction with task-grounded diagnostic labels. A single backbone first predicts an extraction, diagnoses task-specific error labels, and then revises its output conditioned on the diagnosis. Training starts from diagnostic data labeled by an annotation model for cold-start supervised fine-tuning and proceeds through two GRPO stages that reward final extraction quality, format validity, and first-pass correctness, without a process reward model. Experiments on named entity recognition, relation extraction, and event extraction show consistent same-backbone gains over SFT, including 6.83 average F1 on SciER relation extraction, about 20 F1 on out-of-distribution relation extraction, and 14.80 trigger F1 plus 17.50 argument F1 on DuEE1.0. Ablations show that reflection structure is task-sensitive: stronger constraints benefit relation extraction, whereas named entity recognition needs less restrictive correction under domain shift.
\end{abstract}

\section{Introduction}
\label{sec:intro}

Large language models have made rapid progress on open-ended reasoning,from chain-of-thought prompting to reinforcement learning with verifiable rewards,yet a quieter failure persists: they still struggle to reliably extract structured facts from text. When asked to identify entities, relations, or events in a sentence, even capable models produce systematic, recurring errors: missing spans, wrong labels, boundary mismatches, and reversed arguments. Unlike free-form generation, where an imperfect answer can still be useful, information extraction (IE) demands outputs that are simultaneously faithful to the input text and compliant with a predefined schema. This dual constraint makes IE a revealing stress test for structured generation, and a natural setting in which to ask: can a model not only detect that its extraction is wrong, but identify which kind of error it has made, and use that diagnosis to correct itself?

\begin{figure}[t]
\begin{center}
    \includegraphics[width=1\columnwidth]{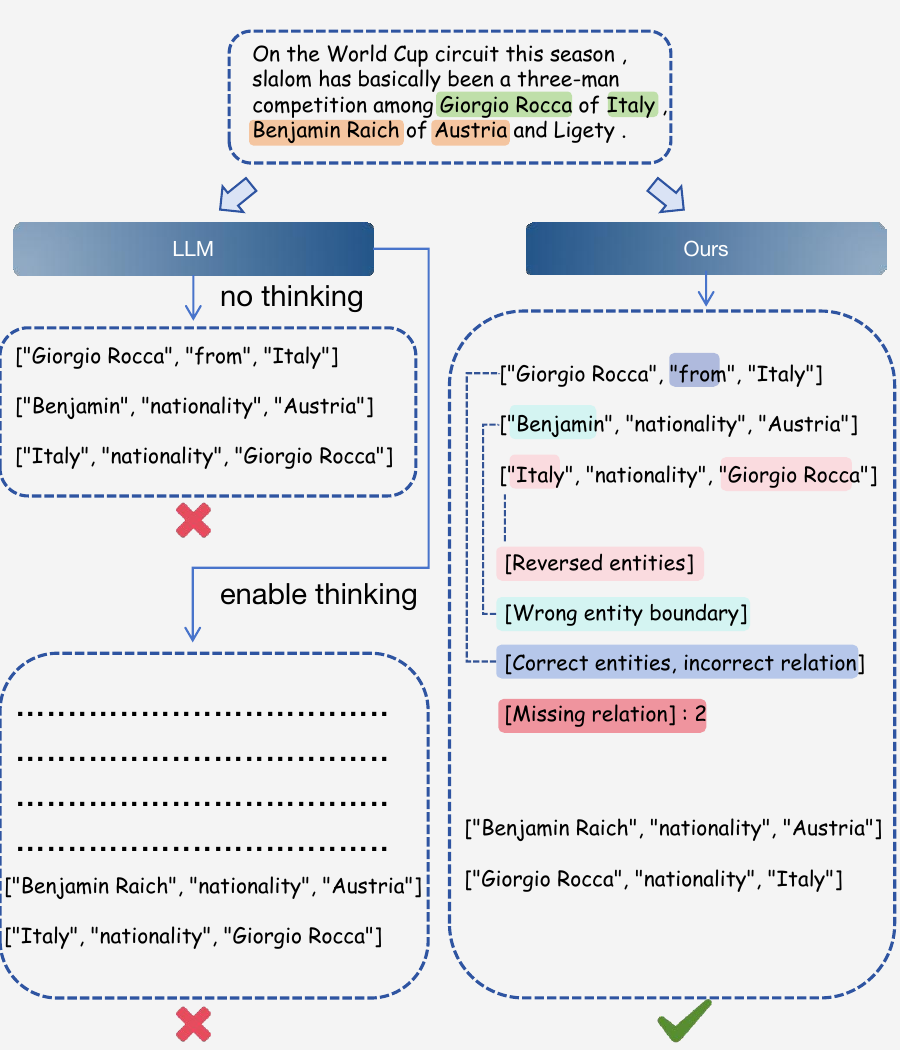}
    \vspace{-2mm}
    \caption{Relation extraction example illustrating why free-form reflection can be insufficient for IE. A standard LLM output contains relation-type, boundary, argument-order, and missing-relation errors; enabling free-form thinking fixes only part of the output. By contrast, \ourmethod{} explicitly diagnoses these errors with task-grounded labels and uses them to guide targeted revision.}
    \label{Fig.intro}
    \vspace{-6mm}
\end{center}
\end{figure}

Reflection and self-correction have been widely studied as mechanisms for improving LLM outputs. Prompting-based methods encourage the model to produce free-form rationales before answering~\citep{wei2022chain,shinn2023reflexion,madaan2023selfrefine} , while training-based approaches use outcome or process rewards to cultivate reasoning behaviors via reinforcement learning ~\citep{lightman2023lets,jaech2024openai,el2025competitive}. Despite their success in mathematical reasoning and code generation, these methods share a common assumption: that reflection is best expressed as unconstrained natural language. For structured IE, this assumption is limiting. A free-form rationale may note that "something is wrong" with an extracted triple, but it rarely pinpoints whether the failure is a missing relation, a wrong entity boundary, or a reversed argument order. Without this diagnostic precision, the model is left to guess at a repair, and the correction step becomes unreliable.

IE offers a distinctive opportunity to move beyond free-form reflection.  
Because its outputs follow explicit schemas, IE errors are not arbitrary  
but systematic and nameable, as shown in ~\citep{zeroshot_IE_ChatGPT,llm-ie-survey-2023}. 
Figure~\ref{Fig.intro} illustrates this concretely.  
A standard LLM produces an invalid relation type (\textit{``from''}),  
truncates \textit{``Benjamin Raich''} to \textit{``Benjamin,''}  
reverses the arguments in [\textit{``Italy''}, \textit{``nationality''},  
\textit{``Giorgio Rocca''}], and misses two nationality relations entirely.  
Enabling free-form chain-of-thought corrects the boundary error but leaves  
the reversed relation untouched.  
The deeper problem is structural: the model senses its output is imperfect,  
yet lacks a mechanism for targeting which error to fix---an invalid type,  
a reversed argument, a boundary mismatch, or a missing relation.  
Diagnostic labels, by contrast, make the repair target explicit:  
{fix the relation type}, {swap the arguments},  
{expand the entity boundary}, {add the missing relations}.  
This observation motivates the central question of our work:  
\textit{Can IE self-correction be organized as label-guided diagnosis  
and revision, rather than as free-form text generation?}

We answer this question with \textbf{LA-RL} (\textbf{L}abel-\textbf{A}ware 
\textbf{R}eflective \textbf{R}einforcement \textbf{L}earning), a framework 
that structures reflection through a discrete, task-grounded diagnostic state. 
Rather than generating open-ended rationales, \textsc{LA-RL} inserts a 
diagnostic step between first-pass extraction and final revision: the model 
first predicts an extraction, diagnoses one or more error labels from a 
predefined task-specific set (e.g., \textit{Missing Relation}, 
\textit{Wrong Entity Boundary}, \textit{Reversed Entities}), and then 
revises its output conditioned on that diagnosis. 
Training proceeds in three stages: (1) cold-start supervised fine-tuning on diagnostic data constructed with an annotation model that labels errors made by a base extractor; (2) a reflection-oriented GRPO stage that rewards final extraction quality and format validity; and (3) a direct-inference enhancement stage that additionally rewards first-pass quality, encouraging the model to solve correctly as early as possible while preserving the diagnostic correction pathway. \textsc{LA-RL} does not train a separate process reward model; all rewards 
are outcome-based and computed directly from extraction metrics.

We evaluate \ourmethod{} on SciER~\citep{zhang2024scier}, OOD RE datasets, and DuEE1.0~\citep{li2020duee} using Qwen2.5-7B-Instruct. \ourmethod{} consistently outperforms same-backbone baselines: +6.83 F1 on SciER RE, $\sim$+20 F1 on OOD RE, and +14.80/+17.50 trigger/argument F1 on DuEE1.0.
Ablations over Generic Reflection, Label-only Correction, and \ourmethod{} show that reflection structure is task-sensitive: RE improves monotonically with stronger constraints, while NER gains no further benefit over label-only correction under domain shift. Reflection structure should thus be matched to task error complexity, not applied uniformly.

\paragraph{Contributions.}  
\begin{itemize}  

\item \textbf{LA-RL}, a label-aware reflective RL framework   
for structured IE self-correction. Rather than relying on   
free-form rationales, LA-RL inserts a discrete diagnostic   
state between first-pass extraction and final revision,   
trained end-to-end with outcome-supervised GRPO.

\item \textbf{A two-stage RL training paradigm.} Beyond the
conventional SFT cold-start followed by a single direct RL
phase, LA-RL decomposes reinforcement learning into
reflection-oriented optimization and direct-inference
enhancement, showing that staged RL yields stable and
cumulative improvements.

\item \textbf{Strong empirical gains.} Under the same   
Qwen2.5-7B-Instruct backbone, LA-RL achieves +6.83 F1 on   
SciER RE, +$\sim$20 F1 on OOD RE, and +14.80/17.50   
trigger/argument F1 on DuEE1.0, outperforming SFT, GRPO,   
and R$^2$GRPO.  

\end{itemize}

\section{Task Formulation}
\label{sec:compo}
Given an input sentence $X = (x_1, x_2, \cdots, x_l)$, the goal of information extraction is to produce a structured output $Y$ that captures task-relevant facts expressed in the text. Depending on the task, $Y$ may denote entity spans and types, relations among entities, or event structures composed of triggers and arguments.

For named entity recognition (NER), the target output is a set of typed spans $Y_{\text{NER}} = \{(e, t)\}$, where $e$ denotes an entity span in the sentence and $t \in \mathcal{T}$ is its entity type. For relation extraction (RE), the target output is a set of structured tuples $Y_{\text{RE}} = \{(e_1, r, e_2)\}$, where $r \in \mathcal{R}$ is a relation type and $e_1, e_2$ are its arguments. For event extraction (EE), the target output is a set of event structures $Y_{\text{EE}} = \{(g, c, A)\}$, where $g$ is an event trigger, $c \in \mathcal{C}$ is an event type, and $A = \{(a_i, \rho_i)\}_{i=1}^{m}$ contains argument spans $a_i$ with their roles $\rho_i$.

We formulate all these IE tasks as constrained text generation. Given a task instruction $\mathcal{I}$ and an input sentence $X$, the model generates a task-specific structured output $Y$ in a predefined format. Under this formulation, different IE tasks share the same high-level inference process, while differing only in their output schema and evaluation criteria.

This unified formulation supports LA-RL: since structured outputs can be checked for fine-grained errors, task-specific diagnostic labels can organize reflection. In our experiments, NER, RE, and EE differ in output formats, error labels, and reward signals.

\section{Label-Aware Self-Reflection for Reinforcement Learning}
\label{sec:rl}
LA-RL instantiates structured reflection by inserting a task-grounded diagnostic state between the first-pass extraction and the revised extraction. The diagnostic state is a discrete set of error labels, such as missing entities, wrong boundaries, missing relations, or reversed arguments, rather than an unrestricted natural-language rationale. We train this behavior in two phases: first constructing label-annotated error data for cold-start SFT, and then applying two-stage outcome-supervised reinforcement learning. The diagnostic labels can be interpreted as a structured reflection space, but they are used here as an intermediate representation inside a standard sequence-generation policy rather than as a new RL problem formulation.

\begin{figure*}[ht]
\centering
\includegraphics[width=0.99\textwidth]{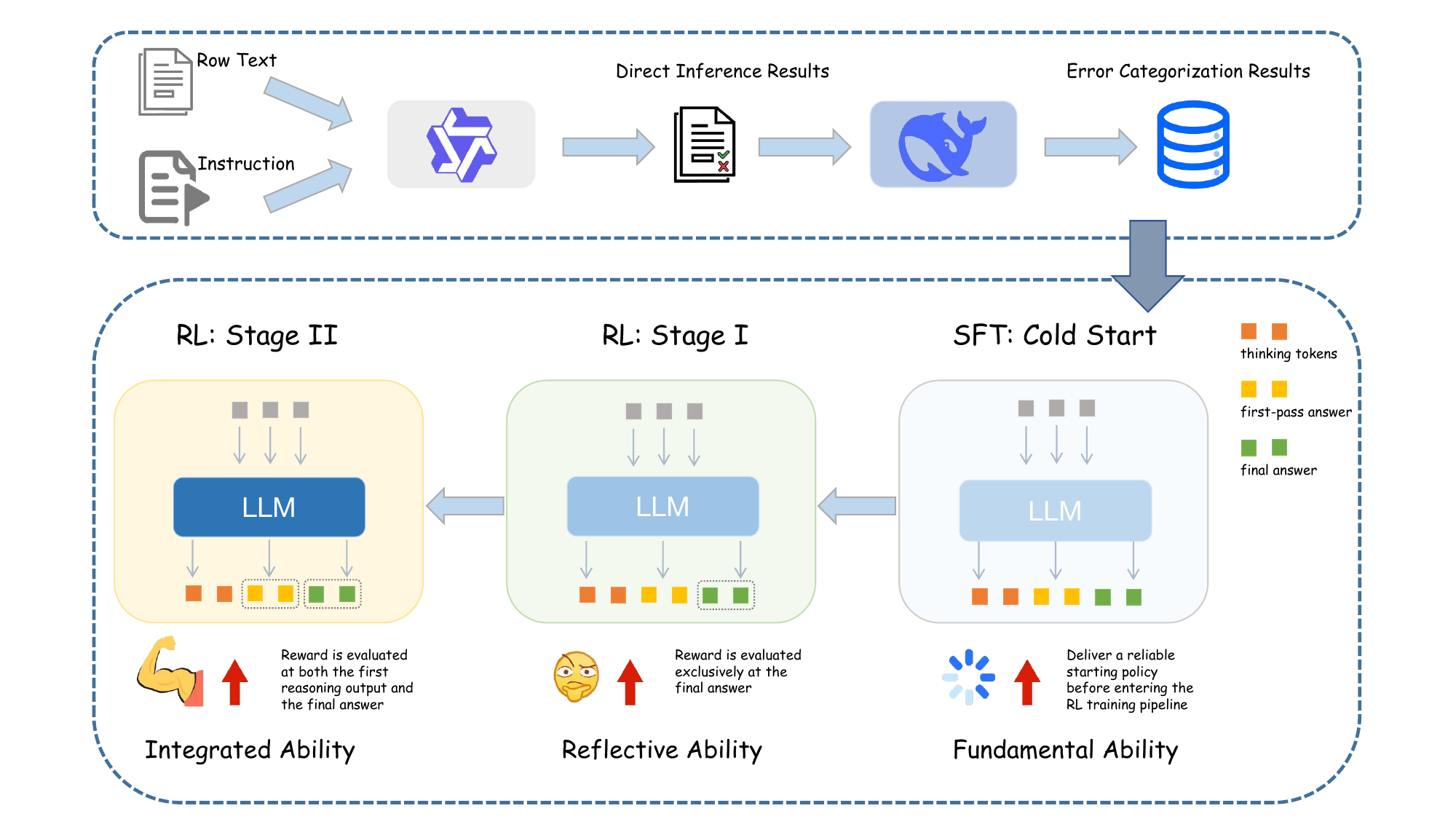}
\caption{Overview of LA-RL. The first-pass extraction is followed by a task-grounded diagnostic state, which structures the reflection step before final revision.}
\label{fig:framework}
\end{figure*}

\subsection{Stage 1: Diagnostic Data Construction}

As summarized in Algorithm~\ref{alg:1}, to construct diagnostic supervision, we use a two-step inference pipeline for structured IE tasks. In the first step, we formulate standardized instruction-style prompts for the target task and apply Qwen2.5-7B-Instruct to perform direct inference on raw texts, obtaining initial structured predictions. As the model is not yet adapted to the task, the generated outputs contain many erroneous instances. These errors provide the raw material for defining task-grounded diagnostic states.

In the second step, we feed each input instance, its gold output, and the corresponding initial prediction into a larger language model, DeepSeek-V3. The annotator compares the initial prediction against the gold output and returns one or more labels from a predefined error set in a structured format. Because a single prediction may contain multiple structural errors simultaneously, we allow the annotator to output an error-label set rather than forcing a single dominant label. The resulting dataset consists of input instances, gold outputs, initial predictions, and diagnostic label sets, which supports cold-start learning of the diagnosis-to-revision pattern.

\renewcommand{\algorithmicrequire}{\textbf{Input:}}
\renewcommand{\algorithmicensure}{\textbf{Output:}}
\begin{algorithm}[t]
    \small
\caption{Diagnostic Error-Label Generation}
    \label{alg:1}
    \begin{algorithmic}[1]
        \Require Training set $\mathcal{D}_{\mathrm{train}}=\{(x_i, y_i)\}_{i=1}^{N}$
        \Statex \hspace{\algorithmicindent} Base IE model $\mathcal{M}_{\mathrm{base}}$, annotator model $\mathcal{M}_{\mathrm{anno}}$
        \Statex \hspace{\algorithmicindent} Error type set $\mathcal{E}$

        \Ensure Diagnostic training set $\mathcal{D}_{\mathrm{diag}}$

        \State Initialize $\mathcal{D}_{\mathrm{diag}} \leftarrow \emptyset$
        \For{each $(x_i, y_i) \in \mathcal{D}_{\mathrm{train}}$}
            \State $\hat{y}_i \leftarrow \mathcal{M}_{\mathrm{base}}(x_i)$ \Comment{Initial prediction}
            \If{$\hat{y}_i \neq y_i$}
                \State $\mathcal{E}_i \leftarrow \mathcal{M}_{\mathrm{anno}}(x_i, y_i, \hat{y}_i, \mathcal{E})$ \Comment{Label set}
                \State $\mathcal{D}_{\mathrm{diag}} \leftarrow \mathcal{D}_{\mathrm{diag}} \cup \{(x_i, y_i, \hat{y}_i, \mathcal{E}_i)\}$
            \EndIf
        \EndFor
        \Return $\mathcal{D}_{\mathrm{diag}}$
    \end{algorithmic}
\end{algorithm}

\subsection{Stage 2: Outcome-Supervised Reflective RL}

After constructing diagnostic supervision, we optimize the policy in two sequential stages that correspond to the ability-oriented blocks in Figure~\ref{fig:framework}. The first stage improves the model's \textit{Reflective Ability}: the model learns to diagnose extraction errors and use diagnostic labels to revise the final answer. The second stage improves the model's \textit{Integrated Ability}: the model learns to combine stronger first-pass extraction with the same diagnostic correction pathway. Both stages use outcome rewards computed from extraction quality and format validity; no process reward model is trained.

\paragraph{Stage I: Improving Reflective Ability.}
We initialize LA-RL from an SFT cold start trained on the diagnostic data. This initialization alleviates reward sparsity and gives the model a basic ability to map extraction errors to one or more diagnostic labels and revise its own predictions under explicit supervision. Starting from this checkpoint, we perform GRPO training with a reward that emphasizes the quality of the final structured answer after diagnostic revision. In this stage, the reward is evaluated exclusively at the final answer: the first reasoning output and the intermediate diagnostic labels are not directly rewarded. Specifically, let $r_{\mathrm{ner}}$ and $r_{\mathrm{re}}$ denote the NER and RE F1 scores of the final prediction, and let $r_{\mathrm{fmt}}$ denote the format reward. We first define the task reward as
\begin{equation}
\label{eq:task_reward}
r_{\mathrm{task}} = \lambda_{\mathrm{ner}} r_{\mathrm{ner}} + \lambda_{\mathrm{re}} r_{\mathrm{re}},
\end{equation}
where $(\lambda_{\mathrm{ner}}, \lambda_{\mathrm{re}}) = (1.0, 2.0)$. The Stage-I reward is then
\begin{equation}
\label{eq:stage1_reward}
r_{\mathrm{I}} = r_{\mathrm{task}} + \lambda_{\mathrm{fmt}} r_{\mathrm{fmt}},
\end{equation}
where $\lambda_{\mathrm{fmt}} = 0.5$. This stage primarily trains the model to use diagnostic reflection to improve the final structured output.

\paragraph{Stage II: Improving Integrated Ability.}
We then continue RL from the Stage-I checkpoint and augment the reward to explicitly value the quality of the first reasoning output, i.e., the first-pass extraction before diagnostic revision. The final-answer reward from Stage I is retained, and an additional first-pass reward is added. Let $r_{\mathrm{re}}^{(1)}$ denote the RE F1 score of the first-pass output. The Stage-II reward is defined as
\begin{equation}
\label{eq:stage2_reward}
r_{\mathrm{II}} = r_{\mathrm{I}} + \lambda_{\mathrm{first}} r_{\mathrm{re}}^{(1)},
\end{equation}
where $\lambda_{\mathrm{first}} = 1.0$. Thus, unlike Stage I, the reward is evaluated at both the first reasoning output and the final answer. This integrated reward encourages the model to solve the task correctly as early as possible, while still preserving the diagnostic reflection step when the initial prediction is imperfect.

\paragraph{Optimization Objective.}
We optimize both stages with Group Relative Policy Optimization (GRPO) \citep{shao2024deepseekmath}, which avoids training a separate value model and fits our outcome-supervised reward design. For each query $x$, we sample a group of outputs $\{o_1, \ldots, o_G\}$ from the old policy $\pi_{\theta_{\mathrm{old}}}$ and compute the corresponding rewards according to Eq.~\eqref{eq:stage1_reward} or Eq.~\eqref{eq:stage2_reward}. Let $\rho_i = \frac{\pi_{\theta}(o_i\mid x)}{\pi_{\theta_{\mathrm{old}}}(o_i\mid x)}$ denote the policy ratio. The policy is updated by maximizing
\begin{align}
    \mathcal{J}(\theta)
    &= \frac{1}{G} \sum_{i=1}^{G}
    \Bigl[
    \min\!\bigl(
    \rho_i A_i,
    \mathrm{clip}(\rho_i, 1-\epsilon, 1+\epsilon) A_i
    \bigr)
    \notag\\
    &\qquad\qquad {}- \beta D_{\mathrm{KL}}\!\left(\pi_{\theta}\,\|\,\pi_{\mathrm{ref}}\right)\Bigr],
    \label{equation:grpo}
\end{align}
where $A_i$ is the normalized group-relative advantage:
\begin{equation}
   A_i = \frac{r_i - \mathrm{mean}(\{r_1, \ldots, r_G\})}{\mathrm{std}(\{r_1, \ldots, r_G\})}.
   \label{equation:grpo_adv}
\end{equation}
A KL regularizer keeps the updated policy close to the reference model from the previous stage, which stabilizes training while allowing the policy to gradually internalize the diagnostic reflection behavior.

\begin{figure*}[ht]
\centering
\includegraphics[width=0.99\textwidth]{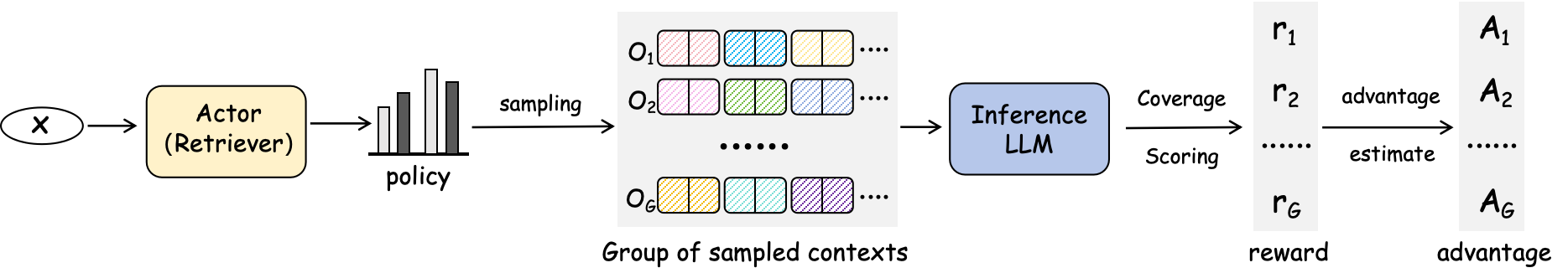}
\caption{Group Relative Policy Optimization (GRPO) process for optimizing the diagnostic reflective policy with outcome rewards.}
\label{fig:grpo_process}
\end{figure*}

\section{Experiments}
\label{sec:exp}

\begin{table*}[!t]
	\centering
	\caption{Test F1 scores of different baselines on SciER and OOD setting. ``Rel'' and ``Rel+'' represent relation extraction under boundary-level and strict evaluation, respectively. The model trained by R$^2$GRPO is used as the starting base model for LA-RL training, and the R$^2$GRPO results are obtained from our reproduction.}
	\vspace{-5pt}
	{\ExpTableStyle
	\begin{tabular}{lcccccc}
		\toprule
		\hline
		\multicolumn{1}{c}{} & \multicolumn{3}{c}{SciER} & \multicolumn{3}{c}{OOD} \\ \cline{2-7}
		\multicolumn{1}{c}{\multirow{-2}{*}{Methods}} & \multicolumn{1}{l}{NER} & \multicolumn{1}{l}{Rel} & \multicolumn{1}{l}{Rel+} & \multicolumn{1}{l}{NER} & \multicolumn{1}{l}{Rel} & \multicolumn{1}{l}{Rel+} \\ \hline
		\multicolumn{7}{c}{\textit{Supervised Baselines}} \\ \hline
		PURE \cite{zhong2020frustratingly} & 81.60 & 53.27 & 52.67 & 71.99 & 50.44 & 49.46 \\
		PL-Marker \cite{ye2021packed} & 83.31 & 60.06 & 59.24 & 73.93 & 59.02 & 56.68 \\
		HGERE \cite{yan2023joint} & \underline{86.85} & \underline{62.32} & \underline{61.10} & \textbf{81.32} & \textbf{61.31} & \textbf{58.32} \\ \hline
		\multicolumn{7}{c}{\textit{Zero-Shot LLMs-based Baselines}} \\ \hline
		Qwen2.5-7B-Instruct & 42.42 & 12.07 & 11.09 & 45.19 & 14.83 & 14.53 \\
		DeepSeek-V3 & 55.81 & 22.63 & 21.94 & 57.98 & 20.92 & 20.22 \\
		DeepSeek-R1 & 71.21 & 28.74 & 28.61 & 67.02 & 28.04 & 27.44 \\
		Kimi-K2-Thinking & 63.12 & 26.58 & 26.35 & 66.69 & 34.22 & 33.91 \\
		Kimi-K2 & 56.21 & 23.22 & 23.07 & 65.69 & 30.03 & 29.66 \\
		Qwen3-Max & 64.91 & 26.29 & 26.14 & 65.92 & 32.61 & 32.64 \\
		Qwen3-Max-Thinking & 65.33 & 28.05 & 27.86 & 65.61 & 31.83 & 31.67 \\
		Gemini-2.5 & 73.33 & 32.01 & 31.67 & 64.61 & 26.32 & 26.07 \\ \hline
		\multicolumn{7}{c}{\textit{Fine-tuned LLMs}} \\ \hline
		GRPO & 70.87 & 37.02 & 34.38 & 59.95 & 27.27 & 23.99 \\
		R$^2$GRPO\cite{li2025pathselectionbetterllms} & 82.81 & 61.49 & 60.30 & 66.70 & 38.80 & 37.76 \\ \hline
		\multicolumn{7}{c}{\textit{Ours}} \\ \hline
		SFT & 82.97 & 59.58 & 57.56 & 65.41 & 36.81 & 35.44 \\
		\StageI & 84.25 & 64.62 & 63.08 & 68.87 & 46.62 & 45.86 \\
		\StageII & \textbf{86.91} & \textbf{66.26} & \textbf{64.53} & \underline{75.34} & \underline{56.68} & \underline{56.03} \\
		\hline
		\bottomrule
	\end{tabular}%
	}
	\label{tab:main_results}
	\vspace{-10pt}
\end{table*}

\subsection{Datasets}
Our evaluation spans multiple representative information extraction settings. For the joint named entity recognition (NER) and relation extraction (RE) setting, we conduct the primary evaluation on the SciER benchmark and its associated out-of-distribution (OOD) datasets \cite{zhang2024scier}. SciER is a widely used scientific IE benchmark comprising 106 research papers annotated with approximately 25k entity mentions and 12k relational instances. To further examine the applicability of the proposed framework beyond the primary NER--RE setting, we additionally evaluate it on DuEE1.0 \cite{li2020duee}, a large-scale Chinese event extraction benchmark constructed from real-world scenarios and comprising approximately 13k instances annotated with 16k events and 33k arguments. Detailed dataset statistics are reported in Tables~\ref{tab:dataset_stat} and \ref{tab:duee_statistics}.

\subsection{Experiment Setup}

\textbf{{Training Settings}}
\textit{{Base Model}}
All our fine-tuning experiments are conducted by adapting the Qwen2.5-7B-Instruct model \cite{qwen2yang2024}.

For the NER/RE experiments, all training settings use Qwen2.5-7B-Instruct as the backbone.
GRPO~\citep{shao2024deepseekmath} denotes direct reinforcement learning with task-level rewards, and R$^2$GRPO~\citep{li2025pathselectionbetterllms} denotes our reproduced Relevance and Rule-Induction Group Relative Policy Optimization baseline.
For our methods, SFT is obtained by performing cold-start SFT on the reproduced R$^2$GRPO-trained model; LA-RL is then initialized from this cold-started model: LA-RL (Stage I) is the intermediate model after the first reinforcement learning stage, and LA-RL (Stage II) is the final model after the second stage.
For the EE experiments, LA-RL is trained directly from Qwen2.5-7B-Instruct rather than from the R$^2$GRPO-trained model.
Following the R$^2$GRPO setup, the RL stages use 1,000 samples selected from the training set for reinforcement learning; the EE experiments follow the same protocol and also use 1,000 sampled training instances.

The system prompts used for joint NER/RE and event extraction (EE) training follow a unified reflection format with task-specific diagnostic labels. We provide the full prompts in Appendix~\ref{sec:appendixPrompt} (Figures~\ref{prompt:system} and~\ref{prompt:ee_system}).

\textbf{{Implementation Details}}
		Detailed hardware, LoRA, SFT/RL hyperparameters, and inference settings are provided in Appendix~\ref{sec:appendixB}.
	
\textbf{{Evaluation Metrics}}
	For Named Entity Recognition (NER) and Relation Extraction (Rel and Rel+), following R$^2$GRPO~\cite{li2025pathselectionbetterllms}, we report the standard micro F1-score. NER: An entity is correct if its span and type match a gold entity. Rel: A relation is correct if the types and spans of both entities and the relation type match a gold relation. Rel+: It further requires the entity type is correct in the triples. For event extraction (EE), we report precision, recall, and F1 for trigger identification (Trig-I), trigger classification (Trig-C), argument identification (Arg-I), and argument classification (Arg-C). Trig-I evaluates trigger span matching, while Trig-C additionally requires the event type to be correct. Arg-I evaluates argument span identification, and Arg-C additionally requires the predicted argument role to match the gold annotation.

\subsection{Results and Analysis}

\paragraph{Main results.}
Table \ref{tab:main_results} reports F1 scores for both NER and RE on the SciER test set and the OOD setting, with RE evaluated under both boundary-level (Rel) and strict (Rel+) settings. We use this table primarily to compare training strategies under the same Qwen2.5-7B-Instruct backbone, while also listing supervised and zero-shot systems for context.

Within the same-backbone fine-tuned setting, LA-RL improves over SFT from 59.58/57.56 to 66.26/64.53 on SciER Rel/Rel+, and from 36.81/35.44 to 56.68/56.03 on OOD Rel/Rel+. It also improves over the pure GRPO baseline trained directly from the base model, which supports the practical role of cold-start SFT before outcome-supervised RL. Compared with Stage I, Stage II further improves relation extraction, suggesting that adding first-pass quality to the outcome reward helps the model preserve direct extraction ability while still using reflection when needed. The substantially larger OOD gain (+19.87 Rel F1) relative to the in-distribution gain (+6.68 Rel F1) suggests that label-aware diagnostic reflection provides stronger generalization benefits under domain shift, where SFT alone is more prone to overfitting to the training distribution. On NER, LA-RL Stage II also improves over SFT by +3.94 F1 on SciER, a gain that is more modest than on RE—a pattern we examine further in the ablation where NER shows a different sensitivity to reflection structure than RE.

LA-RL is competitive with strong supervised systems on SciER and substantially stronger than zero-shot prompting baselines in this setting, but our central conclusion comes from controlled same-backbone comparisons and the structured-reflection ablation in Table~\ref{tab:ablation_reflection}.

\begin{table*}[t]
\centering
\caption{Ablation on structured reflection designs. Generic Reflection removes explicit error-type labels and performs free-form second-pass correction. Label-only Correction uses explicit error labels but removes the full reflective diagnosis process. All variants are evaluated after the second training stage.}
\vspace{-5pt}
{\ExpTableStyle
\begin{tabular}{lcccccc}
\toprule
\multicolumn{1}{c}{} & \multicolumn{3}{c}{SciER} & \multicolumn{3}{c}{OOD} \\ \cline{2-7}
\multicolumn{1}{c}{\multirow{-2}{*}{Methods}} & \multicolumn{1}{l}{NER} & \multicolumn{1}{l}{Rel} & \multicolumn{1}{l}{Rel+} & \multicolumn{1}{l}{NER} & \multicolumn{1}{l}{Rel} & \multicolumn{1}{l}{Rel+} \\ \hline
SFT & 82.97 & 59.58 & 57.56 & 65.41 & 36.81 & 35.44 \\
Generic Reflection & 83.68 & 62.72 & 61.69 & 75.26 & 56.20 & 55.76 \\
Label-only Correction & 83.66 & 64.30 & 62.92 & \textbf{76.21} & 55.78 & 55.33 \\
\ourmethod{} & \textbf{86.91} & \textbf{66.26} & \textbf{64.53} & 75.34 & \textbf{56.68} & \textbf{56.03} \\
\bottomrule
\end{tabular}
}
\label{tab:ablation_reflection}
\vspace{-10pt}
\end{table*}

\paragraph{Ablation on Structured Reflection.}
Table~\ref{tab:ablation_reflection} is the core evidence for the reflection-space structure hypothesis. The variants form a spectrum. Generic Reflection keeps a second-pass correction step but removes explicit diagnostic labels. Label-only Correction exposes error labels but does not require the full diagnosis-to-revision trajectory. LA-RL uses both: the model identifies task-specific diagnostic labels and conditions the final revision on them during training.

The results generally follow this spectrum. On SciER, SFT obtains 59.58/57.56 on Rel/Rel+, Generic Reflection improves to 62.72/61.69, Label-only Correction further improves to 64.30/62.92, and LA-RL reaches 66.26/64.53. On OOD relation extraction, Generic Reflection already brings a large gain over SFT, and LA-RL gives the best Rel/Rel+ scores among the reflection variants. The only exception is OOD NER, where Label-only Correction is slightly higher than LA-RL. We therefore avoid claiming that more structure is uniformly better across every metric; instead, the ablation supports a more modest conclusion that relation extraction improves most consistently when free-form reflection is replaced by a task-grounded diagnosis-to-revision path.

Generic Reflection resembles ordinary self-correction: the second pass may help, but the reflection is not anchored to a specific IE failure mode. Label-only Correction shows that diagnostic labels are useful, but labels alone do not fully capture the benefit. LA-RL performs best on the main relation metrics because it couples the diagnostic state with the revision step under outcome-supervised RL. In this sense, the label set acts as a lightweight way to structure the reflection space, not as a process reward model or a claim about optimal reasoning traces.

\begin{table*}[t]
\centering
\caption{Error-label distribution on the SciER test and OOD relation extraction settings.}
\vspace{-5pt}
{\ExpTableStyle
\begin{adjustbox}{max width=0.98\textwidth,center}
\begin{tabular}{lcccc}
\toprule
\multirow{2}{*}{Error Label} & \multicolumn{2}{c}{SciER} & \multicolumn{2}{c}{OOD} \\ \cline{2-5}
& Count & Ratio & Count & Ratio \\ \hline
Missing relation & 502 & 36.86\% & 130 & 38.24\% \\
Wrong entity/relation types & 418 & 30.69\% & 93 & 27.35\% \\
Wrong entity boundary & 175 & 12.85\% & 41 & 12.06\% \\
Undefined relation & 0 & 0.00\% & 1 & 0.29\% \\
Correct entities, incorrect relation & 258 & 18.94\% & 72 & 21.18\% \\
Reversed entities & 9 & 0.66\% & 3 & 0.88\% \\
\bottomrule
\end{tabular}
\end{adjustbox}
}
\label{tab:error_analysis}
\vspace{-10pt}
\end{table*}

\paragraph{Error Analysis.}
Table~\ref{tab:error_analysis} summarizes the diagnostic error-label distribution on the SciER test and OOD relation extraction settings. Missing relations are the most frequent errors in both settings, accounting for 36.86\% on SciER test and 38.24\% on OOD, followed by wrong entity/relation types. Correct entities with incorrect relations also form a substantial portion of errors, while undefined relations and reversed entities are rare. This distribution suggests that relation extraction failures are dominated by missing relational facts and schema-alignment mistakes, motivating an explicit label-aware diagnosis step before revision.

\begin{table*}[t]
\centering
\caption{Event extraction results on DuEE1.0 across different training stages. Qwen2.5-7B-Instruct denotes zero-shot inference without task-specific training. GRPO denotes pure GRPO training directly from the base model, without SFT cold-start initialization. LA-RL (Stage I) and LA-RL (Stage II) refer to the models obtained after the first and second reinforcement learning stages, respectively. Best results are shown in bold.}
\vspace{-5pt}
{\DenseExpTableStyle
\begin{adjustbox}{max width=0.98\textwidth,center}
\begin{tabular}{lcccccccccccc}
\toprule
\multicolumn{1}{c}{} & \multicolumn{3}{c}{Trig-I} & \multicolumn{3}{c}{Trig-C} & \multicolumn{3}{c}{Arg-I} & \multicolumn{3}{c}{Arg-C} \\ \cline{2-13}
\multicolumn{1}{c}{\multirow{-2}{*}{Method}} & P & R & F1 & P & R & F1 & P & R & F1 & P & R & F1 \\ \hline
Qwen2.5-7B-Instruct
& 52.17 & 33.71 & 40.95
& 45.75 & 29.56 & 35.91
& 21.99 & 16.51 & 18.86
& 19.49 & 14.58 & 16.68 \\
SFT
& 65.39 & 54.80 & 59.63
& 59.97 & 50.25 & 54.68
& 34.42 & 30.04 & 32.08
& 31.62 & 27.59 & 29.47 \\
GRPO
& 63.62 & 61.19 & 62.38
& 58.37 & 56.14 & 57.23
& 33.10 & 35.90 & 34.45
& 30.60 & 33.19 & 31.84 \\
\StageI
& 66.30 & 61.58 & 63.86
& 60.69 & 56.37 & 58.45
& 35.25 & 36.01 & 35.63
& 33.03 & 33.68 & 33.35 \\
\StageII
& \textbf{76.76} & \textbf{72.24} & \textbf{74.43}
& \textbf{74.43} & \textbf{70.05} & \textbf{72.18}
& \textbf{47.68} & \textbf{51.63} & \textbf{49.58}
& \textbf{46.43} & \textbf{50.27} & \textbf{48.27} \\
\bottomrule
\end{tabular}
\end{adjustbox}
}
\label{tab:ee_results}
\vspace{-10pt}
\end{table*}

\paragraph{Generalization to Event Extraction.}
To test whether the same recipe can be instantiated beyond relation extraction, we further evaluate on the DuEE1.0 event extraction benchmark. Event extraction has a different output structure, involving triggers and argument-role assignments, but it still admits task-specific diagnostic labels. As shown in Table~\ref{tab:ee_results}, SFT substantially improves over zero-shot Qwen2.5-7B-Instruct, and LA-RL further improves over SFT across trigger identification, trigger classification, argument identification, and argument classification. Stage II gives the best F1 on all four metric groups. We interpret these results as evidence that the diagnostic-state design is not limited to one relation extraction table, while noting that a broader set of IE tasks would be needed to fully characterize its scope.

Additional qualitative case studies, including a reflection-based correction example and examples where the first-stage prediction is already correct, are provided in Appendix~\ref{sec:appendixD}.


\section{Related Work}
\label{sec:related}
\paragraph{Generative Information Extraction}

Information extraction has traditionally used supervised NER and RE models, such as LSTM-CRF, dependency-tree-based methods, and cross-task architectures, to capture local, syntactic, and interactive signals~\cite{Lample2016,tang2018,9521975}. Recent generative Transformer-based approaches instead formulate IE as text generation, with systems such as Crop, MCL-NER, and zero-shot ChatGPT-based extraction showing the promise of LLMs~\cite{crop,mcl-ner-moying-2023,zeroshot_IE_ChatGPT}. Nevertheless, generative IE remains limited when applied directly without task-specific fine-tuning.

\vspace{-1mm}

\paragraph{Error Samples and Error Labels}

Error samples provide useful supervision because model-generated mistakes often expose systematic weaknesses. Early representation learning and pretraining showed that learning from incorrect or contrasting alternatives can sharpen model discrimination~\cite{mikolov2013distributed,devlin2019bert}. Recent work extends this idea by categorizing LLM hallucinations, integrating error analysis into reasoning, and using related signals in NER, RE, and consistency-based negative-sample selection~\cite{simhi2024distinguishing,lu2023error,10856568,10488145,mo2024c-icl}. Building on these studies, we model recurring IE errors as task-grounded labels, enabling labeled error samples to guide reflection and improve task-specific extraction.

\vspace{-1mm}

\paragraph{RL in LLM}

Reinforcement learning (RL) is a key paradigm for aligning large language models (LLMs). RLHF established the supervised fine-tuning, reward modeling, and PPO pipeline, but remains costly and unstable due to its reliance on human annotations and reward optimization~\cite{ouyang2022training,afzali2024aligning}. Recent variants reduce these costs through AI-generated preferences, direct preference optimization, group-based policy updates, or rule-verifiable rewards~\cite{lee2023rlaif,rafailov2023direct,ding2025multi,guo2025deepseek}.

\section{Conclusion}
\label{sec:conclusion}
This paper studies how to structure reflection for outcome-supervised information extraction. Instead of unconstrained long rationales, LA-RL places a task-grounded discrete diagnostic state between first-pass extraction and final revision, combining annotation-model diagnostic labels, cold-start SFT, and two-stage GRPO with outcome rewards without a separate process reward model.

Experiments show consistent same-backbone gains over SFT and pure GRPO. Ablations indicate that relation extraction benefits from structured diagnosis-to-revision rather than free-form self-correction, and DuEE1.0 results show the same recipe can extend to event extraction, though broader task coverage is needed for stronger generalization claims. Overall, reflection-space structure is an important design dimension for reflective LLM training.

\section{Limitations}
\label{sec:limitation}
Despite its contributions, this study has the following limitations: (1) although we evaluate LA-RL on representative IE tasks including NER, RE, and EE, broader settings such as document-level extraction, low-resource IE, and multilingual IE remain underexplored; (2) LA-RL currently relies on manually defined diagnostic labels, and future work can explore automatic methods for generating diverse sets of error-label categories to facilitate transfer across different IE tasks; (3) our experiments validate LA-RL using only a 7B-parameter backbone model, and whether the method remains similarly effective when scaled to larger backbone models requires further exploration.



\bibliography{custom}

\appendix
\onecolumn

\section{Dataset Statistics}
\label{sec:appendixA}
For relation extraction, we use the SciER benchmark as the primary dataset together with several out-of-distribution test sets; their detailed entity and relation distributions are summarized in Table~\ref{tab:dataset_stat}.
\begin{table}[h!]
		\centering
		\caption{Statistics of the SciER benchmark and OOD test sets used for relation extraction}
		\vspace{-5pt}
		{\ExpTableStyle
			\begin{tabular}{lccccc}
				\toprule
				\textbf{Entity/Relation Type} & \textbf{Train} & \textbf{Dev} & \textbf{SciER Test} & \textbf{OOD Test} & \textbf{Total} \\ \hline
				\multicolumn{6}{c}{\textit{Entity Types}} \\ \hline
				\texttt{Method} & 11424 & 1549 & 1890 & 1018 & 15881 \\
				\texttt{DATASET} & 3220 & 269 & 370 & 83 & 3942 \\
				\texttt{TASK} & 3397 & 416 & 688 & 194 & 4695 \\
				\textbf{Total} & 18041 & 2234 & 2948 & 1295 & 24518 \\ \hline
				\multicolumn{6}{c}{\textit{Relation Types}} \\ \hline
				\texttt{PART-OF} & 1865 & 214 & 304 & 111 & 2494 \\
				\texttt{USED-FOR} & 2398 & 343 & 546 & 167 & 3454 \\
				\texttt{EVALUATED-WITH} & 863 & 78 & 131 & 49 & 1121 \\
				\texttt{SYNONYM-OF} & 880 & 76 & 170 & 89 & 1215 \\
				\texttt{COMPARE-WITH} & 875 & 175 & 114 & 54 & 1218 \\
				\texttt{SUBCLASS-OF} & 697 & 114 & 176 & 73 & 1060 \\
				\texttt{BENCHMARK-FOR} & 551 & 64 & 85 & 28 & 728 \\
				\texttt{SUBTASK-OF} & 210 & 31 & 65 & 9 & 315 \\
				\texttt{TRAINED-WITH} & 404 & 37 & 35 & 2 & 478 \\
				\textbf{Total} & 8743 & 1132 & 1626 & 582 & 12083 \\
				\bottomrule
			\end{tabular}%
		}
		\label{tab:dataset_stat}
		\vspace{-10pt}
		\end{table}

For event extraction, we use the DuEE1.0 benchmark; its detailed split statistics are summarized in Table~\ref{tab:duee_statistics}.

\begin{table}[h!]
		\centering
		\caption{Statistics of the DuEE1.0 benchmark used for event extraction. Since the test sets for DuEE1.0 are not open-sourced, we use the valid sets as our evaluation data.}
		\vspace{-5pt}
		{\ExpTableStyle
			\begin{tabular}{lccc}
				\toprule
				\textbf{Metric} & \textbf{Train} & \textbf{Dev} & \textbf{Total} \\ \hline
				Instances & 11908 & 1492 & 13400 \\
				Events & 13860 & 1783 & 15643 \\
				Arguments & 28905 & 3682 & 32587 \\
				\bottomrule
			\end{tabular}%
		}
		\label{tab:duee_statistics}
		\vspace{-10pt}
		\end{table}

\section{Implementation Experiment Details}
\label{sec:appendixB}
All experiments are conducted using the PyTorch deep learning framework on a single NVIDIA A800 GPU.
To ensure stable training and decoding, we carefully select the experimental settings and hyperparameters.
The inference-stage hyperparameters used throughout all experiments are summarized in Table~\ref{tab:inference_parameters}, the supervised fine-tuning (SFT) hyperparameters are summarized in Table~\ref{tab:sft_parameters}, and the reinforcement learning (RL) hyperparameters are summarized in Table~\ref{tab:rl_parameters}.
\begin{table}[!t]
\centering
\caption{Inference-stage hyperparameters used throughout all experiments.}
\vspace{-5pt}
{\ExpTableStyle
\begin{tabular}{lc}
\toprule
Inference-stage parameter & Value \\ \hline
Max sequence length & 8192 \\
Top-$p$ & 1.0 \\
Top-$k$ & 1 \\
Temperature & 0 \\
\bottomrule
\end{tabular}%
}
\label{tab:inference_parameters}
\vspace{-10pt}
\end{table}
\begin{table}[!t]
\centering
\caption{SFT-stage hyperparameters used in LA-RL training.}
\vspace{-5pt}
{\ExpTableStyle
\begin{tabular}{lc}
\toprule
SFT-stage parameter & Value \\ \hline
\texttt{learning\_rate} & $1 \times 10^{-4}$ \\
\texttt{optim} & \texttt{adamw\_8bit} \\
\texttt{weight\_decay} & 0.01 \\
\texttt{lr\_scheduler\_type} & linear \\
\texttt{num\_train\_epochs} & 2 \\
\texttt{per\_device\_train\_batch\_size} & 1 \\
\texttt{gradient\_accumulation\_steps} & 8 \\
\texttt{lora\_rank} & 16 \\
\texttt{lora\_alpha} & 32 \\
\bottomrule
\end{tabular}%
}
\label{tab:sft_parameters}
\vspace{-10pt}
\end{table}
\begin{table}[!t]
\centering
\caption{RL-stage hyperparameters used in LA-RL training.}
\vspace{-5pt}
{\ExpTableStyle
\begin{tabular}{lc}
\toprule
RL-stage parameter & Value \\ \hline
\texttt{learning\_rate} & $5 \times 10^{-6}$ \\
\texttt{adam\_beta1} & 0.9 \\
\texttt{adam\_beta2} & 0.99 \\
\texttt{weight\_decay} & 0.1 \\
\texttt{warmup\_ratio} & 0.1 \\
\texttt{temperature} & 1.0 \\
\texttt{lr\_scheduler\_type} & cosine \\
\texttt{optim} & \texttt{adamw\_8bit} \\
\texttt{num\_train\_epochs} & 2 \\
\texttt{max\_grad\_norm} & 0.1 \\
\texttt{lora\_rank} & 64 \\
\texttt{lora\_alpha} & 128 \\
\texttt{per\_device\_train\_batch\_size} & 16 \\
\texttt{num\_generations} & 16 \\
\texttt{gradient\_accumulation\_steps} & 1 \\
\bottomrule
\end{tabular}%
}
\label{tab:rl_parameters}
\vspace{-10pt}
\end{table}

For the inference stage, we use a maximum sequence length of 8192 with deterministic decoding, setting top-$p$ to 1.0, top-$k$ to 1, and temperature to 0 throughout all experiments in this paper. This deterministic setup reduces sampling noise and helps ensure stable and reproducible evaluation.

For the SFT stage, we fine-tune the model for two epochs using 8-bit AdamW with a linear learning-rate scheduler, a learning rate of $1 \times 10^{-4}$, and weight decay of 0.01. We use LoRA adapters with rank 16 and scaling factor 32, set the per-device training batch size to 1, and accumulate gradients for 8 steps before each optimizer update.

For the RL stage, we optimize the policy for two epochs using 8-bit AdamW with cosine learning-rate decay, a warmup ratio of 0.1, a sampling temperature of 1.0, and LoRA adapters with rank 64 and scaling factor 128. We use a per-device batch size of 16, sample 16 generations for each prompt during policy optimization, and set the gradient accumulation steps to 1. The standalone GRPO baseline in the main paper uses the same RL-stage hyperparameters, but it is trained directly from Qwen2.5-7B-Instruct and does not use the SFT checkpoint as a cold start.

For event extraction, we adopt the same two-stage RL design as in the NER/RE setting, but use event-specific evaluation signals. Let $r_{\mathrm{trig}}$ and $r_{\mathrm{arg}}$ denote the trigger and argument F1 scores of the final prediction, respectively, and let $r_{\mathrm{fmt}}$ denote the format reward. We first define the EE task reward as
\begin{equation}
\label{eq:ee_task_reward}
r_{\mathrm{task}}^{\mathrm{EE}} = \lambda_{\mathrm{trig}} r_{\mathrm{trig}} + \lambda_{\mathrm{arg}} r_{\mathrm{arg}},
\end{equation}
where $(\lambda_{\mathrm{trig}}, \lambda_{\mathrm{arg}}) = (1.0, 2.0)$, since argument prediction is the more difficult and structurally informative part of EE.

The Stage-I reward is then
\begin{equation}
\label{eq:ee_stage1_reward}
r_{\mathrm{I}}^{\mathrm{EE}} = r_{\mathrm{task}}^{\mathrm{EE}} + \lambda_{\mathrm{fmt}} r_{\mathrm{fmt}},
\end{equation}
where $\lambda_{\mathrm{fmt}} = 0.5$.

In Stage II, we further encourage stronger direct inference by adding the first-pass argument F1. Let $r_{\mathrm{arg}}^{(1)}$ denote the argument F1 score of the first-pass output. The Stage-II reward is defined as
\begin{equation}
\label{eq:ee_stage2_reward}
r_{\mathrm{II}}^{\mathrm{EE}} = r_{\mathrm{I}}^{\mathrm{EE}} + \lambda_{\mathrm{first}} r_{\mathrm{arg}}^{(1)},
\end{equation}
where $\lambda_{\mathrm{first}} = 1.0$.

\section{Supplementary Experiments}
\label{sec:appendixSupplementaryExperiments}
\paragraph{Zero-shot Evaluation on Unseen RE Datasets.}
We include a small zero-shot evaluation in which the model is trained on SciER and directly evaluated, without target-domain fine-tuning, on CoNLL04 and NYT10. Table~\ref{tab:performance_comparison} shows that LA-RL outperforms the Qwen2.5-7B-Instruct baseline across both datasets. We treat this result as a supporting observation that the learned reflection behavior can transfer to some unseen RE datasets, while leaving broader cross-domain generalization to future work.

\begin{table}[H]
\centering
\caption{Zero-shot relation extraction performance on unseen datasets. Best results are shown in bold.}
\vspace{-5pt}
{\ExpTableStyle
\begin{adjustbox}{max width=0.98\linewidth,center}
\begin{tabular}{lcc}
\toprule
Model & CoNLL04 & NYT10 \\ \hline
Qwen2.5-7B-Instruct & 0.3379 & 0.1159 \\
LA-RL & \textbf{0.3550} & \textbf{0.1347} \\
\bottomrule
\end{tabular}
\end{adjustbox}
}
\label{tab:performance_comparison}
\vspace{-10pt}
\end{table}

\section{Prompt}
\label{sec:appendixPrompt}
We provide the system prompts used for task-specific reflective training. Figure~\ref{prompt:system} shows the prompt for joint named entity recognition (NER) and relation extraction (RE), and Figure~\ref{prompt:ee_system} shows the corresponding prompt for event extraction (EE).

\begin{figure}[ht]
\begin{tcolorbox}[
	colback=gray!10,
	colframe=gray!80,
	title=System Prompt,
	fonttitle=\bfseries,
	boxrule=0.3mm,
	sharp corners,
	width=\textwidth
	]
	\noindent Respond in the following format:\\
	<think>\\
Provide step-by-step reasoning to solve the task based on the given instructions and sentence.
Because the joint NER and RE extraction task is difficult, you need to try one derivation in your thinking and output the result in JSON format as \texttt{\{"ner": [[entity, entity type], ...], "rel": [[subject, relation, object], ...]\}}.
The reasoning process must include a reflection step: if an error occurs, categorize it under one or more common error labels for the current task and revise your reasoning accordingly. For NER, use [Missing entities], [Wrong entity types], [Wrong entity boundary], [Undefined entity types], or [Spurious entities]. For RE, use [Missing relation], [Wrong entity/relation types], [Wrong entity boundary], [Undefined relation], [Correct entities, incorrect relation], or [Reversed entities]. If no error is found, explicitly return an empty label set \{\}.\\
	</think>\\
	<answer>\\
	Provide the final answer in JSON format as specified in the instruction.\\
	</answer>
	\vspace{-4pt}
\end{tcolorbox}
\caption{System prompt used in LA-RL training for joint named entity recognition and relation extraction with task-specific NER and RE error labels.}
\label{prompt:system}
\end{figure}

\begin{figure}[ht]
\begin{tcolorbox}[
	colback=gray!10,
	colframe=gray!80,
	title=System Prompt,
	fonttitle=\bfseries,
	boxrule=0.3mm,
	sharp corners,
	width=\textwidth
	]
	\noindent Respond in the following format:\\
	<think>\\
Provide step-by-step reasoning to solve the task based on the given instructions and sentence.
Because the EE task is difficult, you need to try one derivation in your thinking and output the result.
The reasoning process must include a reflection step: if an error occurs, categorize it under one of the common error labels ([Correct trigger, incorrect arguments], [Missing trigger], [Spurious trigger], [Wrong trigger boundary], [Wrong event type], [Missing argument], [Spurious argument], [Wrong argument boundary], [Wrong argument role], [Argument attached to wrong event]) and revise your reasoning accordingly. If no error is found, explicitly return an empty label set \{\}.\\
	</think>\\
	<answer>\\
	Provide the final answer in JSON format as specified in the instruction.\\
	</answer>
	\vspace{-4pt}
\end{tcolorbox}
\caption{System prompt used in LA-RL training for event extraction with task-specific EE error labels.}
\label{prompt:ee_system}
\end{figure}


\section{Label Definition Explanation}
\label{sec:appendixC}
\begin{figure*}[ht]
\begin{center}
    \centering
    \includegraphics[width=0.9\columnwidth]{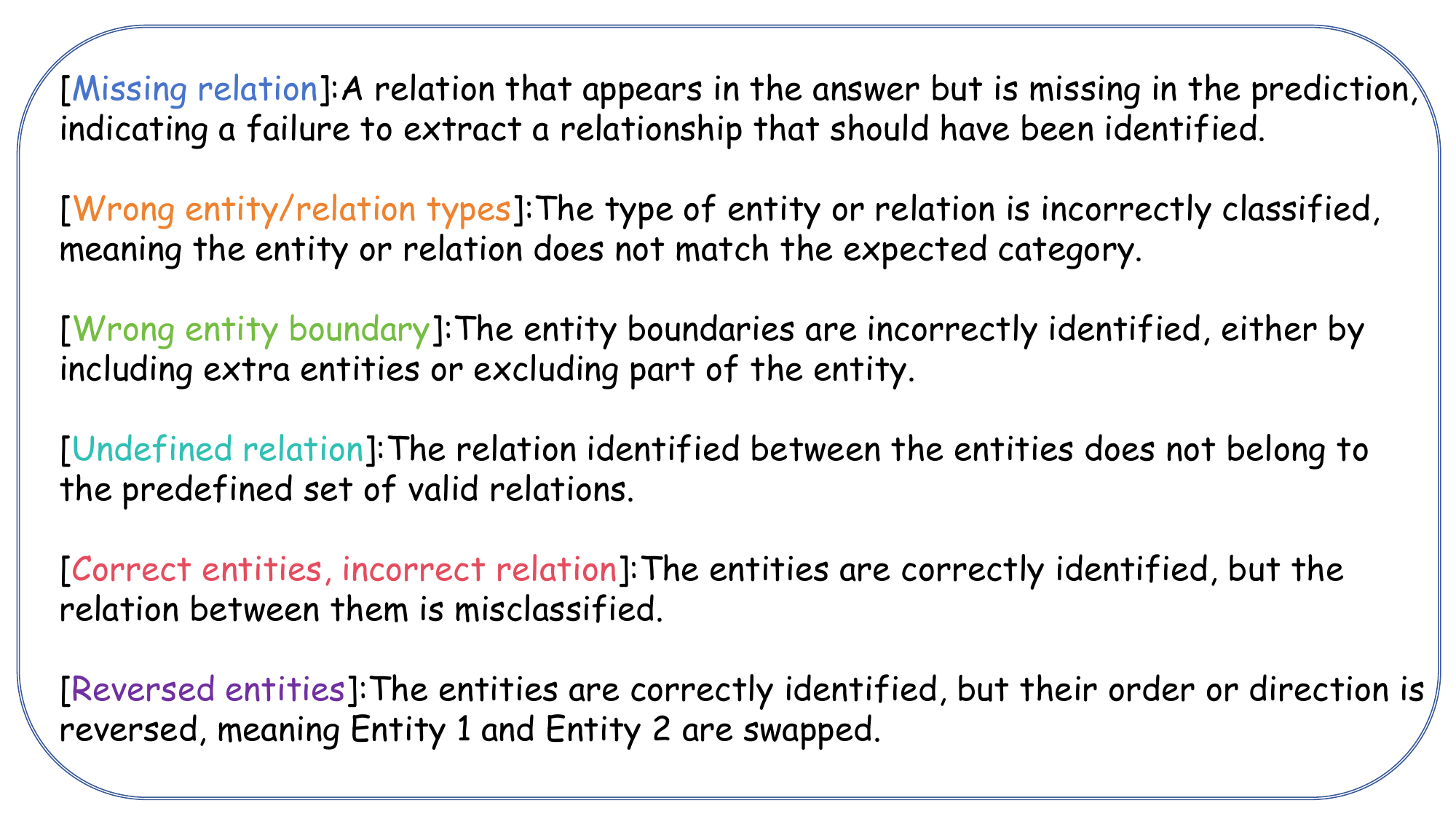}
 
    \vspace{-3mm}
    \caption{Relation extraction error label set used for multi-label diagnosis.
The RE setting uses a six-label set.}
    \label{Fig.label_def} 
    \vspace{-5mm}
\end{center}
\end{figure*}

Figure \ref{Fig.label_def} illustrates the relation extraction error label set and the corresponding definitions used for multi-label diagnosis in our method.
For the Relation Extraction (RE) task, we use a six-label set for diagnosing common error sources: \textit{Missing relation}, \textit{Wrong entity/relation types}, \textit{Wrong entity boundary}, \textit{Undefined relation}, \textit{Correct entities, incorrect relation}, and \textit{Reversed entities}. When multiple structural errors co-occur in a single prediction, the diagnosis may contain more than one label from this set.
Each label in the set is immediately followed by an explanation of its definition.

Table~\ref{tab:neg_label_distribution} reports the empirical distribution of individual label assignments in the RE multi-label diagnosis dataset. \textit{Missing relation} is the most frequently assigned label, accounting for more than half of all label assignments.

\begin{table}[ht]
\centering
\caption{Distribution of individual label assignments in the RE multi-label diagnosis dataset. Ratios are rounded to two decimals.}
\vspace{-5pt}
{\ExpTableStyle
\begin{tabular}{lrr}
\toprule
RE error label & Count & Ratio \\ \hline
Missing relation & 6975 & 50.83\% \\
Wrong entity/relation types & 3140 & 22.88\% \\
Wrong entity boundary & 1560 & 11.37\% \\
Correct entities, incorrect relation & 1295 & 9.44\% \\
Undefined relation & 460 & 3.35\% \\
Reversed entities & 293 & 2.14\% \\
\bottomrule
\end{tabular}%
}
\label{tab:neg_label_distribution}
\vspace{-10pt}
\end{table}

For the Event Extraction (EE) task, Table~\ref{tab:ee_neg_label_definitions} summarizes the ten-label set used for multi-label diagnosis of trigger- and argument-level errors.

\begin{table}[H]
\centering
\caption{Event extraction error label set used in our multi-label diagnosis framework.}
\vspace{-5pt}
{%
\scriptsize
\setlength{\tabcolsep}{4.4pt}%
\renewcommand{\arraystretch}{1.00}%
\begin{adjustbox}{max width=0.985\linewidth,center}
\begin{tabular}{@{}ll@{}}
\toprule
EE error label & Definition \\ \hline
\textit{Missing trigger} & A gold event trigger is not predicted by the model. \\
\textit{Spurious trigger} & The model predicts a trigger that does not correspond to any gold event mention. \\
\textit{Wrong trigger boundary} & The trigger is detected, but its span boundary is incorrect. \\
\textit{Wrong event type} & The trigger span is matched, but the assigned event type is incorrect. \\
\textit{Correct trigger, incorrect arguments} & The trigger prediction is correct, but the associated argument set is wrong. \\
\textit{Missing argument} & A gold argument of an event is omitted. \\
\textit{Spurious argument} & The model predicts an extra argument that is not supported by the gold annotation. \\
\textit{Wrong argument boundary} & The argument role is predicted on the wrong span boundary. \\
\textit{Wrong argument role} & The argument span is identified, but its semantic role is assigned incorrectly. \\
\textit{Argument attached to wrong event} & An argument is predicted for an incorrect trigger or linked to the wrong event instance. \\
\bottomrule
\end{tabular}%
\end{adjustbox}
}
\label{tab:ee_neg_label_definitions}
\vspace{-10pt}
\end{table}

\section{Case Study}
\label{sec:appendixD}
In this appendix, we provide additional qualitative case studies for LA-RL, covering both reflection-based correction and cases in which the first-stage prediction is already correct.

\subsection{Reflection-Based Correction}
\begin{figure}[H]
\centering
\includegraphics[width=\linewidth]{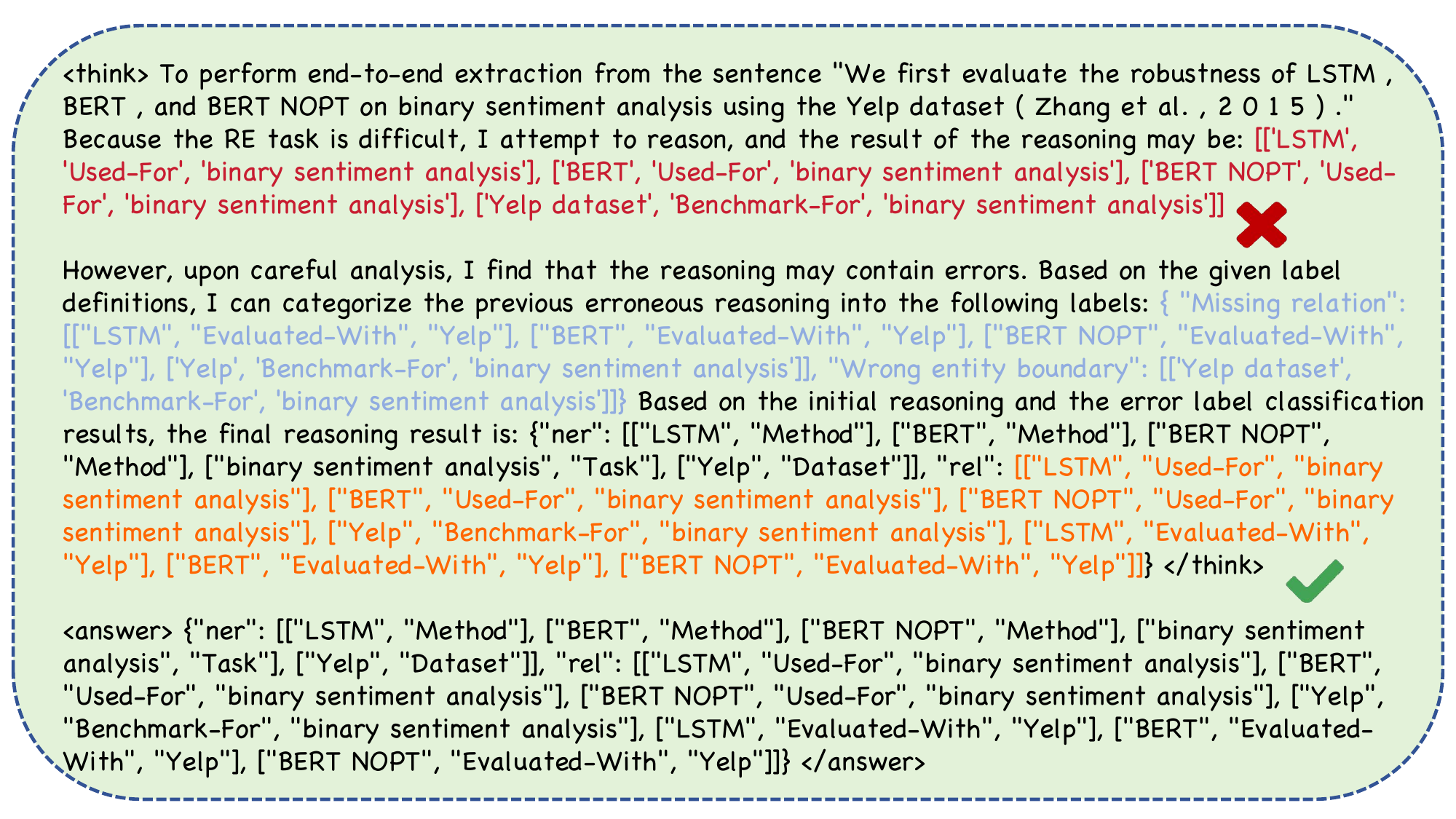}
\vspace{-3mm}
\caption{A representative case in which label-aware reflection corrects missing relations and entity-boundary errors.}
\label{Fig.example1}
\vspace{-3mm}
\end{figure}

Figure~\ref{Fig.example1} presents a representative example in which the initial prediction contains structured errors that are corrected during reflection.

We use the following input sentence:

``We first evaluate the robustness of LSTM, BERT, and BERT NOPT on binary sentiment analysis using the Yelp dataset (Zhang et al., 2015).''

This sentence contains multiple entity types, including methods (LSTM, BERT, BERT NOPT), a task (binary sentiment analysis), and a dataset (Yelp), together with several implicit semantic relations.

In the first-stage inference, the model produces the following initial reasoning result:

``the result of the reasoning may be: [['LSTM', 'Used-For', 'binary sentiment analysis'], ['BERT', 'Used-For', 'binary sentiment analysis'], ['BERT NOPT', 'Used-For', 'binary sentiment analysis'], ['Yelp dataset', 'Benchmark-For', 'binary sentiment analysis']]''

Although the model correctly identifies major entities and some relations, this output exhibits two common errors. First, it fails to extract the implicit Evaluated-With relations between methods and the dataset, corresponding to a Missing relation error. Second, it uses ``Yelp dataset'' instead of the canonical entity ``Yelp,'' resulting in a Wrong entity boundary error.

In the second stage, the model performs label-aware reflection by explicitly diagnosing and categorizing these errors, as indicated by:

``I find that the reasoning may contain errors... {'Missing relation': ..., 'Wrong entity boundary': ...}''

Guided by the identified error types, the model revises its reasoning and generates the final structured output:

``ner'': [["LSTM", "Method"], ["BERT", "Method"], ["BERT NOPT", "Method"], ["binary sentiment analysis", "Task"], ["Yelp", "Dataset"]]

``rel'': [["LSTM", "Used-For", "binary sentiment analysis"], ["BERT", "Used-For", "binary sentiment analysis"], ["BERT NOPT", "Used-For", "binary sentiment analysis"], ["Yelp", "Benchmark-For", "binary sentiment analysis"], ["LSTM", "Evaluated-With", "Yelp"], ["BERT", "Evaluated-With", "Yelp"], ["BERT NOPT", "Evaluated-With", "Yelp"]]

Compared with the initial inference, the final result recovers all missing Evaluated-With relations and normalizes the dataset entity boundary. This example illustrates how the two-stage label-aware reflective mechanism enables effective self-correction and leads to more complete and consistent extraction results.

\subsection{Correct First-Stage Inference}
\begin{figure}[h]
\centering
\includegraphics[width=0.9\linewidth]{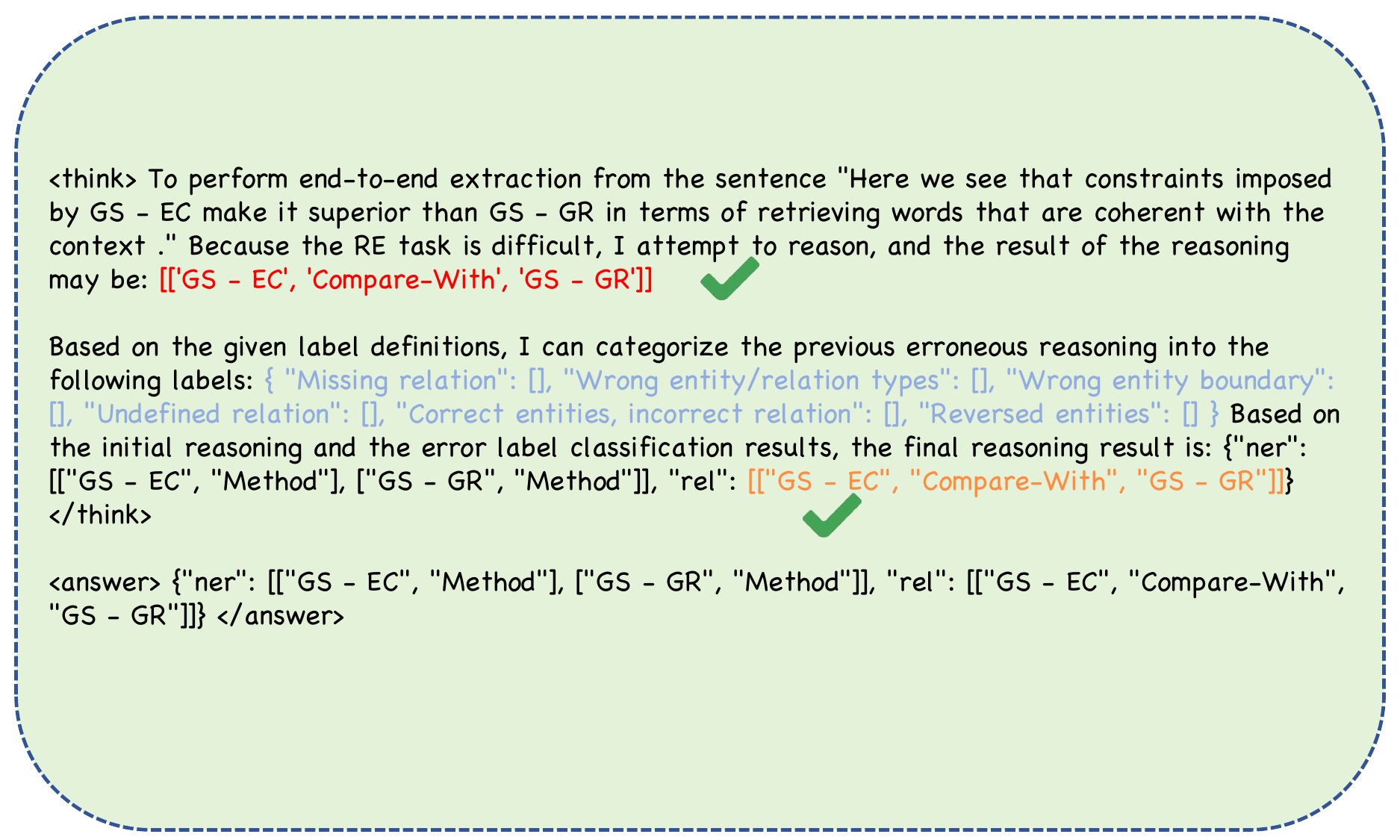}
\caption{An example of correct direct inference in the first stage.}
\label{fig:other_case_study}
\end{figure}

Figure~\ref{fig:other_case_study} shows an additional example in which the first-stage direct inference is already correct.

This case study analyzes a scenario in which the proposed two-stage label-aware reflective framework produces a correct extraction result during the first-stage direct inference, demonstrating that the reflection mechanism does not introduce unnecessary modifications when the initial reasoning is accurate.

The input sentence is:

``Here we see that constraints imposed by GS - EC make it superior than GS - GR in terms of retrieving words that are coherent with the context.''

This sentence describes a comparative relationship between two methods, GS - EC and GS - GR, explicitly indicating their relative performance.

During the first-stage direct inference, the model generates the following reasoning result:

``the result of the reasoning may be: [['GS - EC', 'Compare-With', 'GS - GR']]''

The model correctly identifies both entities as Method types and assigns the appropriate semantic relation Compare-With. Importantly, the subsequent error categorization step reports no detected errors across all predefined categories, including Missing relation, Wrong entity boundary, and Incorrect relation type, indicating high confidence in the initial prediction.

Based on this assessment, the model preserves the original reasoning and outputs the final structured result as:

``ner'': [["GS - EC", "Method"], ["GS - GR", "Method"]]

``rel'': [["GS - EC", "Compare-With", "GS - GR"]]

This case highlights an important property of the proposed framework: the reflection stage acts as a conditional correction mechanism rather than an obligatory revision step. When the direct inference result is already consistent with label definitions and relation semantics, the model maintains the original output without introducing spurious changes.

Overall, this example demonstrates that the two-stage label-aware reflective reinforcement learning framework is both corrective and conservative. It not only enables effective self-correction in the presence of errors but also preserves correct predictions, ensuring robustness and stability in end-to-end information extraction tasks.

\clearpage

\end{document}